%% file: samplepaper.tex
\documentclass[runningheads]{llncs}
\usepackage{graphicx}
\usepackage{microtype}
\usepackage{graphicx}
\usepackage{url}
\usepackage{algorithm}
\usepackage{algorithmic}
\usepackage{subfigure}
\usepackage{booktabs} 
\usepackage{multirow}
\usepackage{multicol}
\usepackage{amsfonts, amsmath}
\usepackage{xcolor}
\usepackage{soul}
\usepackage{adjustbox}
\usepackage{amssymb}
\usepackage{pifont}
\newcommand{\xmark}{\ding{55}}%
\newcommand{\cmark}{\ding{51}}
\usepackage{enumitem}

\begin{document}
\title{Federated Learning Meets Fairness and Differential Privacy}
%
%
\author{Manisha Padala, Sankarshan Damle and Sujit Gujar}
\institute{Machine Learning Lab,\\ International Institute of Information Technology (IIIT), Hyderabad
\email{\{manisha.padala,sankarshan.damle\}@research.iiit.ac.in}
\email{sujit.gujar@iiit.ac.in}
}
\authorrunning{Padala, Damle and Gujar}
%

%
\maketitle              
\begin{abstract}
Deep learning's unprecedented success raises several ethical concerns ranging from biased predictions to data privacy. Researchers tackle these issues by introducing fairness metrics, or federated learning, or differential privacy. A first, this work presents an ethical federated learning model, incorporating all three measures simultaneously. Experiments on the Adult, Bank and Dutch datasets highlight the resulting ``empirical interplay" between accuracy, fairness, and privacy.

\end{abstract}
\section{Introduction}

Deep Learning's success is made possible in part due to the availability of big datasets -- distributed across several owners. To resolve this, researchers propose \emph{Federated Learning}~(FL), which enables parallel training of a unified model~\cite{mcmahan2017communication}. Such a requirement naturally arises for mobile phones, network sensors, and other IoT applications. In FL, these respective owners are referred to as `clients' (henceforth \textit{agents}). The agents individually train a model on their private data.  A `central server,' referred henceforth as an \emph{aggregator}, receives the individual models and computes a single overall model through different heuristics for achieving high performance on any test data~\cite{wahab2021federated}.

The data available with each agent is often imbalanced or biased. Machine learning models may further amplify the bias present. More concretely, when trained only for achieving high accuracy, the model predictions become highly biased towards certain demographic groups like gender, age, or race \cite{barocas16,berk,chouldechova17}. Such groups are known as the \emph{sensitive attributes}. Post the impossibility result on achieving a perfectly unbiased model~\cite{chouldechova17}, researchers propose several approaches which focus on minimizing the bias while maintaining high accuracy~\cite{zafar17,madras18,agarwal18,manisha2018fnnc}. 

Invariably all these approaches require the knowledge of the sensitive attribute. These attributes often comprise the most critical information. The law regulations at various places prohibit using such attributes to develop ML models. E.g., the EU General Data Protection Regulation prevents the collection of sensitive user attributes~\cite{tran2020differentially}. Thus, it is imperative to address discrimination while preserving the leakage of sensitive attributes from the data samples.

Observing that the aggregator in FL has no direct access to private data or sensitive attributes, prima facie preserves privacy. However, there exist several attacks that highlight the information leak in an FL setting~\cite{mothukuri2021survey}. To plug this information leak, researchers either use cryptographic solutions based mainly on complex \emph{Partial Homomorphic Encryption} (PHE) or use Differential Privacy (DP). Private FL solutions using PHE (e,g.,~\cite{mohassel2017secureml,yang2019federated,zhang2020batchcrypt,fang2021privacy}) suffer from computational inefficiency and post-processing attacks. Thus, in this work we focus on the strong privacy guarantees provided by a \emph{differentially private} solution~\cite{pathak2010multiparty,shokri2015privacy,naseri2020toward,ieeefpdp}.

\smallskip
\noindent \emph{FPFL Framework.} A first, we incorporate both fairness and privacy guarantees for an FL setting through our novel framework \emph{FPFL}: Fair and Private Federated Learning. Our primary goal is to simultaneously preserve the privacy of the training data and the sensitive attribute while ensuring fairness. With FPFL, we achieve this goal by ingeniously decoupling the training process in two \textit{phases}. In Phase 1, each agent trains a model, on its private dataset, for fair predictions. Then in Phase 2, the agents train a differentially private model to \textit{mimic} the fair predictions from the previous model. At last, each agent communicates the private model from Phase 2 to the aggregator. 

\smallskip
\noindent \emph{Fairness and Privacy Notions.}
The deliberate phase-wise training ensures an overall fair and accurate model which does not encode any information related to the sensitive attribute. Our framework is general and absorbs any fairness or DP metrics. It also allows any fairness guaranteeing technique in Phase 1. In this paper we demonstrate FPFL's efficacy w.r.t. the state-of-the-art technique and the following notions.
\begin{enumerate}
    \item Fairness: We consider \emph{demographic parity} (DemP) and \emph{equalized odds} (EO). DemP states that a model's predictions are independent of a sensitive attribute of the dataset~\cite{dwork2012fairness}. EO states that the \emph{false positive rates} and \emph{false negative rates} of a model are equal across different groups or independent of the sensitive attribute~\cite{hardt16}. \item Privacy: We quantify the privacy guarantees within the notion of $(\epsilon,\delta)$-\emph{local differential privacy}~\cite{dwork2014algorithmic}. The notion of local-DP is a natural fit for FL. In our setting, the aggregator acts as an adversary with access to each agent's model. We show that with local-DP, the privacy of each agent's training data and sensitive attribute is protected from the aggregator.
\end{enumerate}


\smallskip
\noindent\textit{Empirical interplay between accuracy, fairness, and privacy.} The authors in \cite{bagdasaryan2019differential} were one of the first to show that ensuring privacy may come at a cost to fairness. While the trade-off between fairness and privacy in ML is under-explored, Table~\ref{tab:RW} presents the existing literature. Apart from those, the authors in \cite{li2020} consider \emph{accuracy parity} which is weaker than EO in a non-private FL setting.

\smallskip
\noindent\textit{Contributions.} In summary, we propose our novel FPFL framework (Fig.~\ref{fig::FPFL-framework}). We prove that FPFL provides the local-DP guarantee for both the training data and the sensitive attribute(s) (Proposition~\ref{claim::FPFL}). Our experiments on the Adult, Bank, and Dutch datasets show the empirical trade-off between fairness, privacy, and accuracy of an FL model (Section~\ref{sec::exp}).

\input{relatedwork}
\section{Preliminaries}
We consider a binary classification problem with $\mathcal{X}$ as our ($d$-dimensional) instance space, $\mathcal{X} \in \mathbb{R}^d$; and our output space as $\mathcal{Y} \in \{0,1\}$. We consider a single sensitive attribute $\mathcal{A}$ associated with each individual instance. Such an attribute may represent sensitive information like age, gender or caste. Each $a \in \mathcal{A}$ represents a particular category of the sensitive attribute like male or female. 

\paragraph{Federated Learning Model.}
Federated Learning (FL) decentralizes the classical machine learning training process. FL comprises two type of actors: (i) a set of agents $\mathbb{A}=\{1,\dots, m\}$ where each agent $i$ owns a \emph{private} dataset $\mathcal{X}_i$\footnote{Let $|\mathcal{X}_i|$ denote the cardinality of $\mathcal{X}_i$ with $\mathbf{X}=\sum_i |\mathcal{X}_i|$.}\footnote{We use the sub-script ``$i$" when referring to a particular agent $i$ and drop it when not referring to any particular agent.}; and (ii) an \emph{Aggregator}. Each agent provides its model, trained on its dataset, to the aggregator. The aggregator's job then is to derive an overall model, which is then communicated back to the agents. This back-and-forth process continues until a model with sufficient accuracy is derived.

At the start of an FL training, the aggregator communicates an initial, often random, set of model parameters to the agents. Let us refer to the initial parameters as $\theta_0$. At each timestep $t$ each agent updates their individual parameters denoted by $\theta_{(i,t)}$, using their private datasets. The agents then communicate the updated parameters to the aggregator, who derives an overall model through different heuristics~\cite{wahab2021federated}. In this paper, we focus on the weighted sum heuristics, i.e., the overall model parameters take the form $\theta_t=\sum_{j\in \mathbb{A}}\frac{|\mathcal{X}_j|}{\mathbf{X}}\cdot \theta_{(j,t)}$. To distinguish the final overall model, we refer to it as $\theta^*$, calculated at a timestep $T$.

\paragraph{Fairness Metrics.}
We consider the following two fairness constraints. 
\begin{definition}[Demographic Parity (DemP)] \label{def:demp} A classifier $h$ satisfies Demographic Parity under a distribution over $(\mathcal{X}, \mathcal{A}, \mathcal{Y})$ if its predictions $h(\mathcal{X})$ is independent of the sensitive attribute $\mathcal{A}$. That is, $\forall a \in \mathcal{A} \ and \ p \in \{0,1\}$,
$$ \Pr[h(\mathcal{X})=p | \mathcal{A} = a] = \Pr[h(\mathcal{X})=p]$$
Given that $p \in \{0,1\}$, we have $\forall a$ $$\mathbb{E}[h(\mathcal{X}) | \mathcal{A} = a] = \mathbb{E}[h(\mathcal{X})]. $$ 
\end{definition}

\begin{definition}[Equalized Odds (EO)] \label{def:eo}A classifier $h$ satisfies Equalized Odds under a distribution over $(\mathcal{X}, \mathcal{A}, \mathcal{Y})$ if its predictions $h(\mathcal{X})$ are independent of the sensitive attribute $\mathcal{A}$ given the label $\mathcal{Y}$. That is, $\forall a \in \mathcal{A}, \ p \in \{0,1\} \ and \ y \in \mathcal{Y}$
$$ 
\Pr[h(\mathcal{X})=p | \mathcal{A} = a, \mathcal{Y} = y] = \Pr[h(\mathcal{X})=p | \mathcal{Y}=y]
$$
Given that $p \in \{0,1\}$, we can say $\forall a, y$ 
$$
\mathbb{E}[h(\mathcal{X}) | \mathcal{A} = a, \mathcal{Y}=y] = \mathbb{E}[h(\mathcal{X}) | \mathcal{Y}=y].
$$ 
\end{definition}



\paragraph{Local Differential Privacy (LDP).} We now define LDP in the context of our FL model. We remark that LDP does not require defining adjacency.

\begin{definition}[Local Differential 
Privacy (LDP)~\cite{dwork2014algorithmic}]\label{def:ldp}
For an input set $\mathcal{X}$ and the set of noisy outputs $\mathcal{Y}$, a randomized algorithm $\mathcal{M}:\mathcal{X}\rightarrow \mathcal{Y}$ is said to be $(\epsilon,\delta)$-LDP if $\forall x, x'\in \mathcal{X}$ and $\forall y\in\mathcal{Y}$ the following holds,
    \begin{equation}\label{eqn::LDP}
        \Pr[\mathcal{M}(x)=y] \leq \exp(\epsilon)\Pr[\mathcal{M}(x')=y] + \delta.
    \end{equation}
\end{definition}

LDP provides a statistical guarantee against an inference which the adversary can make based on the output of $\mathcal{M}$. This guarantee is upper-bounded by $\epsilon$, referred to as the \emph{privacy budget}. $\epsilon$ is a metric of \emph{privacy loss} defined as,
\begin{equation}\label{eqn::PL}
    L^y_{\mathcal{M}(x)||\mathcal{M}(x')} = \ln\left(\frac{\mathcal{M}(x)=y}{\mathcal{M}(x')=y}\right).
\end{equation}

The privacy budget, $\epsilon$, controls the trade-off between quality (or, in our case, the accuracy) of the output vis-a-vis the privacy guarantee. That is, there is no ``free-dinner" -- lower the budget, better the privacy but at the cost of quality. The ``$\delta$" parameter in \eqref{eqn::LDP} allows for the violation of the upper-bound $\epsilon$, but with a small probability.

Differentially private ML solutions focus on preserving an individual's privacy within a dataset. Such privacy may be compromised during the training process or based on the predictions of the trained model~\cite{fredrikson2015model}. The most famous of such an approach is the DP-SGD algorithm, introduced in \cite{abadi2016deep}. In DP-SGD, the authors sanitize the gradients provided by the Stochastic Gradient Descent (SGD) algorithm with \emph{Gaussian} noise ($\mathcal{N}(0,\sigma^2)$). This step aims at controlling the impact of the training data in the training process. 



\paragraph{Adversary Model.} Towards designing a private FL system, it suffices to provide DP guarantees for any possible information leak to the aggregator. The post-processing properties of DP further preserves the DP guarantee for the training data and the sensitive attribute from any other party. 

We consider the ``black-box" model for our adversary, i.e., the aggregator has access to the trained model and can interact with it via inputs and outputs. With this, it can perform model-inversion attacks~\cite{fredrikson2015model}, among others.




\section{FPFL: Fair and Private Federated Learning}
\begin{figure}[!t]
    \centering
    \begin{minipage}{.5\textwidth}
        \centering
        \includegraphics[width=\columnwidth]{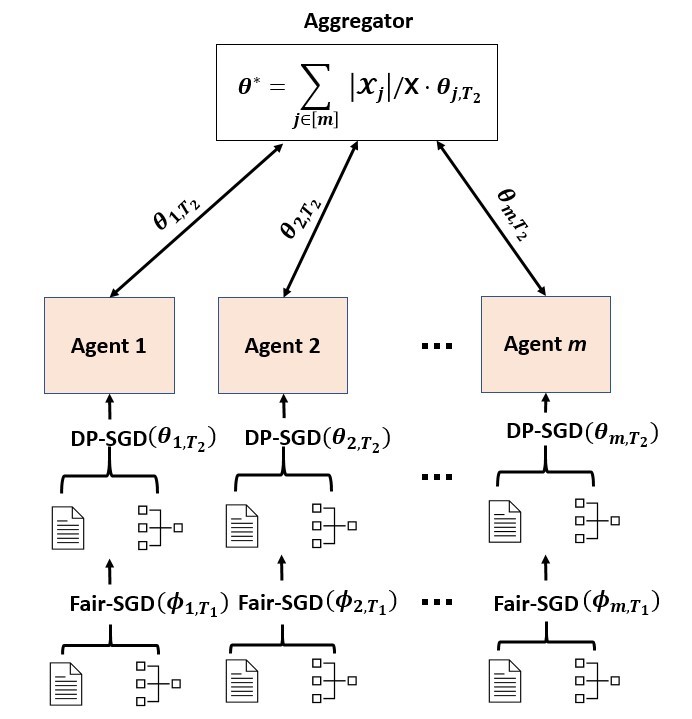}
        \caption{FPFL Model.}
        \label{fig::FedML}
    \end{minipage}%
\hfill
    \fbox{
        {
    \begin{minipage}{0.45\textwidth}
    
        \begin{center}
        \textbf{FPFL Framework}
    \end{center}
    
    \begin{enumerate}[noitemsep, leftmargin=*]
        \item \textbf{Initialization}
        \item \textbf{Local Training Process.} Each agent $i\in\mathbb{A}$, invokes Algorithm~\ref{algo::P1}
        
        \item \textbf{Local Training Process.} Each agent $i\in\mathbb{A}$, invokes Algorithm~\ref{algo::P2}
        \item Local training process ends
        \item \textbf{Model Aggregation.} Aggregator computes and then broadcasts an overall model
        \item Agents re-initialize their local models with the overall model received
        \item Repeat steps 3-6 till a sufficient overall accuracy is reached (Eq.~\ref{eq:overall})
    \end{enumerate}
       \caption{FPFL Framework}
         \label{fig::FPFL-framework}
    \end{minipage}}}
\end{figure}

In FPFL (Figure~\ref{fig::FedML}), we consider a classification problem. Each agent $i$ deploys two multi-layer \emph{neural networks} (NNs) to learn the model parameters in each phase. The training comprises of \textit{two} phases: (i) In Phase 1, each agent privately trains a model on its private dataset to learn a highly fair and accurate model; and (ii) In Phase 2, each agent trains a second model to mimic the first, with DP guarantees. This process is akin to knowledge distillation \cite{hinton2015distilling}. In FPFL, only the model trained in Phase 2 is broadcasted to the aggregator. 

To enhance readability and to remain consistent with FL notations, we denote the model parameters learned for Phase 1 with $\phi$ and Phase 2 with $\theta$. Likewise, we represent the total number of training steps in Phase 1 with $T_1$, and for Phase 2, we use $T_2$. 

%
     
%

\smallskip
\noindent\textbf{Phase 1: Fair-SGD.}
In this phase, we train the network to maximize accuracy while achieving the best possible fairness on each agent's private dataset. We adapt the \emph{Lagrangian Multiplier method}~\cite{manisha2018fnnc} to achieve a fair and accurate model. We denote the model for agent $i$ as $h^{\phi_i}$ with parameters $\phi_i$. Briefly, the method trains a network with a unified loss that has two components. The first component of the loss maximizes accuracy, i.e., the cross-entropy loss, 
\begin{equation*}
        l_{CE}(h^{\phi_i}, \mathcal{X}, \mathcal{Y}) =\underset{(x,y)\sim (\mathcal{X}, \mathcal{Y})}{\mathbb{E}}[-y_i\log(h^{\phi_i}(x)) - (1-y)\log(1- h^{\phi_i}(x))]
\end{equation*}

The second component of the loss is a specific fairness measure. For achieving DemP (Definition \ref{def:demp}), the loss function is given by,
\begin{equation}
\label{eq:dp}
l_{DemP}(h^{\phi_i},\mathcal{X},\mathcal{A}) = |\mathbb{E}[h^{\phi_i}(x) | \mathcal{A} = a] - \mathbb{E}[h^{\phi_i}(x)]| 
\end{equation} 
For achieving EO (Definition 
\ref{def:eo}), the corresponding loss function is,
\begin{equation}
\label{eq:eo}
    l_{EO}(h^{\phi_i},\mathcal{X},\mathcal{A},\mathcal{Y}) = |\mathbb{E}[h^{\phi_i}(x) | \mathcal{A} = a, y] - \mathbb{E}[h^{\phi_i}(x)|y]|
\end{equation}
Hence, the overall loss from the Lagrangian method is, 
\begin{equation}\label{eqn::Phase1Loss}
\begin{split}
    L_{1}(h^{\phi_i}, \mathcal{X}, \mathcal{A}, Y) = l_{CE} +\lambda l_k, \quad k \in \{DemP, EO\}
\end{split}{}
\end{equation}
In the equation above, $\lambda \in \mathbb{R}^+$ is the Lagrangian multiplier. The overall optimization is as follows: $\underset{\phi}{min}\  \underset{\lambda}{max} \ L_{1}(\cdot)$.
Thus, each agent trains the Fair-SGD model $h_i^{\phi}$ to obtain the best accuracy w.r.t. a given fairness metric. We present it formally  in~Algorithm~\ref{algo::P1}.

\smallskip
\noindent\textbf{Phase 2: DP-SGD.}
In this phase, the agents train a model that is communicated with the aggregator. This model denoted by $h^{\theta_i}$ is trained by each agent $i$ to learn the predictions of its own Fair-SGD model ($h^{\phi_i}$) from Phase 1. The loss function is given by, 
\begin{equation}\label{eqn::Phase2Loss}
\begin{split}
        L_2(h^{\theta_i}, h^{\phi_i}) =\underset{x \sim \mathcal{X}}{\mathbb{E}}[ - h^{\phi_i}(x) \log(h^{\theta_i}(x)) -
         (1- h^{\phi_i}(x))\log(1- h^{\phi_i}(x))]
    \end{split}
\end{equation}
Equation \ref{eqn::Phase2Loss} is the cross-entropy loss between predictions from DP-SGD model and the labels given by the predictions from Fair-SGD model. That is, $L_2(\cdot)\downarrow \implies \theta_i \rightarrow \phi_i.  $

To preserve privacy of training data and sensitive attribute, we use $(\epsilon,\delta)$-LDP (Definition \ref{def:ldp}). In particular, we deploy  DP-SGD. In it, the privacy of the training data is preserved by sanitizing the gradients provided by SGD with \emph{Gaussian} noise ($\mathcal{N}(0,\sigma^2)$). Given that the learnt model $h^{\theta_i}$, mimics $h^{\phi_i}$, the learnt model is reasonably fair and accurate. Algorithm~\ref{algo::P2} formally presents the training.

\input{FPFL-Algo}

\smallskip
\noindent\textbf{FPFL Framework.}
The $\theta_i$'s from each agent are communicated to the aggregator for further performance improvement. The aggregator takes a weighted sum of the individual $\theta_i$'s and broadcasts it to the agents. The agents further train on top of the aggregated model before sending it to the aggregator. This process gets repeated to achieve the following overall objective,
\begin{equation}\label{eq:overall}
    \theta^* = \underset{\theta}{\mbox{arg min}} \sum_{j \in [m]}\frac{|\mathcal{X}_j|}{\mathbf{X}} \cdot L_2(h^{\theta}, h^{\phi_j}).
\end{equation}

We now formally couple these processes above to present the FPFL framework with Figure~\ref{fig::FPFL-framework}. The framework presents itself as a \textit{plug-and-play} system, i.e., a user can use any other loss function instead of $L_1, L_2$, or change the underlying algorithms for fairness and DP, or do any other tweak it so desires.

\smallskip

\noindent\textbf{FPFL: Differential Privacy Bounds. }
Observe that the model learned in Phase 1, $h^\phi$, requires access to both the training data ($\mathcal{X}$) and the sensitive attribute ($\mathcal{A}$). Fortunately, this phase is entirely independent of the FL aggregation process. In contrast, the model learned in Phase 2, $h^\theta$ -- trained to mimic the predictions of $h^\phi$ -- is communicated to the aggregator. 

Any information leak in FPFL may take place in the following two ways. Firstly, training data may get compromised through  $h^\theta$. Secondly, mimicking the predictions from $h^\phi$ may, in turn, leak information about the sensitive attribute.
We observe that the DP guarantee for the training data follows from~\cite[Theorem 1]{abadi2016deep} directly. The following proposition proves that the training process in Phase 2 does not leak any additional information regarding $\mathcal{A}$ to the aggregator. Then, Corollary~\ref{corollary::FPFL} uses the result with \cite[Theorem 1]{abadi2016deep} to provide the privacy bounds.


\begin{proposition}\label{claim::FPFL}
With the differentially private FPFL framework~(Figure~\ref{fig::FPFL-framework}), the aggregator with access to the model $h^\theta$ learns no additional information, over the DP guarantee, regarding the sensitive attribute $\mathcal{A}$.
\end{proposition}


\begin{corollary}\label{corollary::FPFL}
For the FPFL framework~(Figure~\ref{fig::FPFL-framework}), $\forall i\in\mathbb{A}$ there exists constants $c_1$ and $c_2$, with the sampling probability $q_i=B_i/\mathcal{X}_i$ and the total number of timesteps $T$ in Phase 2, such that for any $\epsilon_i<c_1q_i^2T$, the framework satisfies $(\epsilon_i,\delta_i)$-LDP for $\delta_i>0$ and for
    $$
     \sigma_i \geq c_2\frac{q_i\sqrt{T\ln(1/\delta_i)}}{\epsilon_i}.
    $$
\end{corollary}

\section{FPFL: Experimental Results\label{sec::exp}}

\paragraph{Datasets.}  We conduct experiments on the following three datasets: Adult~\cite{manisha2018fnnc}, Bank~\cite{manisha2018fnnc} and Dutch~\cite{dutch}. The first two have $\approx 40k$ samples, while the Dutch dataset has $\approx 60k$. In the Adult dataset, the task is a binary prediction of whether an individual's income is above or below USD 50000. The sensitive attribute is \textit{gender} and is available as either male or female. In the Bank dataset, the task is to predict if an agent has subscribed to the term deposit or not. In this case, we consider \textit{age} as the sensitive attribute. We consider only two categories for age, where people between the ages 25 to 60 form the majority group, and those under 25 or over 60 form the minority group. In the Dutch dataset, similar to Adult, we consider \textit{gender} as the sensitive attribute comprising males and females. The task is to predict the occupation. For training an FL model, we split the datasets such that each agent has an equal number of samples. In order to do so, we duplicate the samples in the existing data -- especially the minority group -- to get exactly $50k$ samples for the first two datasets. Despite this, each agent ends up with an uneven distribution of samples belonging to each attribute maintaining the data heterogeneity. This results in heterogeneous data distribution among the agents. We hold $20\%$ of the data from each dataset as the test set.

\paragraph{Hyperparameters.} For each agent, we train two fully connected neural networks having the same architecture. Each network has two hidden layers with $(500, 100)$ neurons and ReLU activation. For DemP, we consider 5 agents in our experiments and split datasets accordingly. To estimate EO, we need sufficient samples for both sensitive groups such that each group has enough samples with both the possible outcomes. In the Adult dataset, we find only $3\%$ female samples earning above USD 50000. Similarly, in the Bank dataset, the minority group that has subscribed to the term deposit forms only $1\%$ of the entire data. Due to this, in our experiments for EO, we consider only 2 agents.

\paragraph{Training Fair-SGD (Phase 1).} We use Algorithm \ref{algo::P1}, with $\eta=0.001$ and $B=500$. The optimizer used is Adam for updating the loss using the Lagrangian multiplier method. For the Adult dataset, we initialize with $\lambda = 10$, and for the Bank and Dutch datasets, we initialize with $\lambda = 5$. The model is trained for $200$ epochs.

\paragraph{Training DP-SGD (Phase 2).} We use Algorithm \ref{algo::P2}, with $\eta=0.25$, $B=500$, and the clipping norm $C = 1.5$. For the optimizer we use the Tensorflow-privacy library's Keras DP-SGD optimizer\footnote{\url{https://github.com/tensorflow/privacy}}. We train the model in this phase for 5 epochs locally before aggregation. This process is repeated 4 times, i.e. $T = 20$. 

\paragraph{Baselines.} To compare the resultant interplay between accuracy, fairness, and privacy in FPFL, we create the following two baselines. 
\begin{itemize}
    \item[B1] In this, the agents train the model in an FL setting only for maximizing accuracy without any fairness constraints in the loss. 
    \item[B2] To obtain B2, each agent trains the model in an FL setting for both accuracy and fairness using Algorithm~\ref{algo::P1} with DemP loss \eqref{eq:dp} or EO loss \eqref{eq:eo}.
\end{itemize}
For both B1 and B2, the final model obtained after multiple aggregations is used to report the results. These baselines maximize accuracy and ensure fairness without any privacy guarantee. Essentially, this lack of a privacy guarantee implies that for both baselines, we skip FPFL's Phase 2.

\paragraph{$(\epsilon,\delta)$-bounds.} We calculate the bounds for an agent from Corollary~\ref{corollary::FPFL}.
To remain consistent with the broad DP-ML literature, we vary $\epsilon$ in the range $(0,10]$ by appropriately selecting $\sigma$ (noise multiplier). Observe that $\epsilon\rightarrow\infty$ for B1 and B2 as the sensitivity is unbounded. As standard $\forall i \in \mathbb{A}$, we keep $\delta=10^{-4}<1/|\mathcal{X}_i|$ for DemP and $\delta=0.5\times 10^{-4}<1/|\mathcal{X}_i|$ for EO.

\paragraph{DemP and EO.} When the loss for DemP (\ref{eq:dp}) and EO (\ref{eq:eo}) is exactly zero, the model is perfectly fair. As perfect fairness is impossible, we try to minimize the loss. In our results, to quantify the fairness guarantees, we plot $l_{DemP}$ and $l_{EO}$ on the test set. \textit{Lower} the values, \textit{better} is the guarantee. For readability we refer $l_{DemP}$ and $l_{EO}$ as DemP and EO in our results.

%
\begin{figure*}[!t]
\centering
\subfigure[Demographic Parity (DemP)]{\includegraphics[width=0.4\columnwidth]{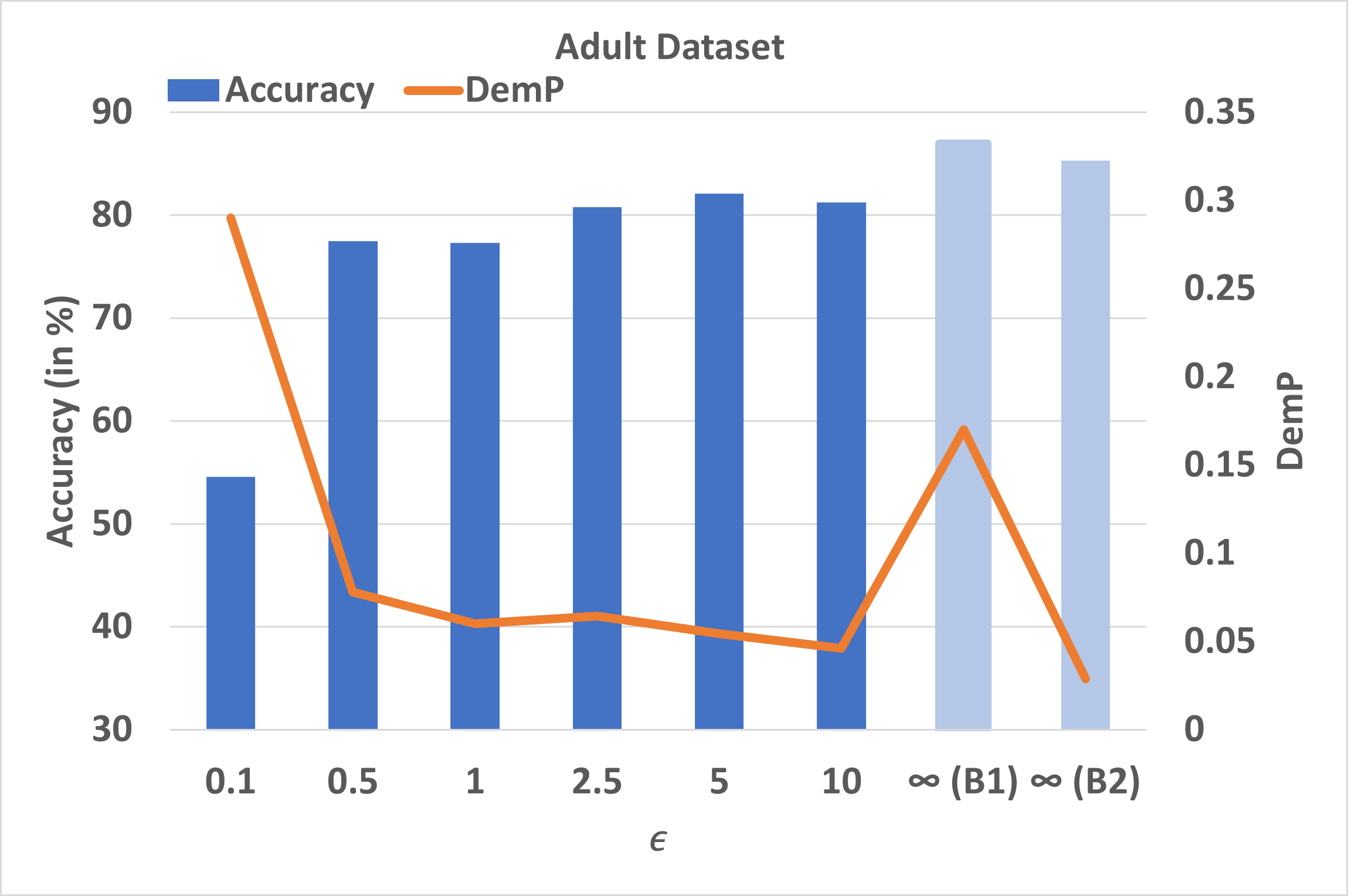}}
\smallskip
\subfigure[Equalized Odds (EO)]{\includegraphics[width=0.4\columnwidth]{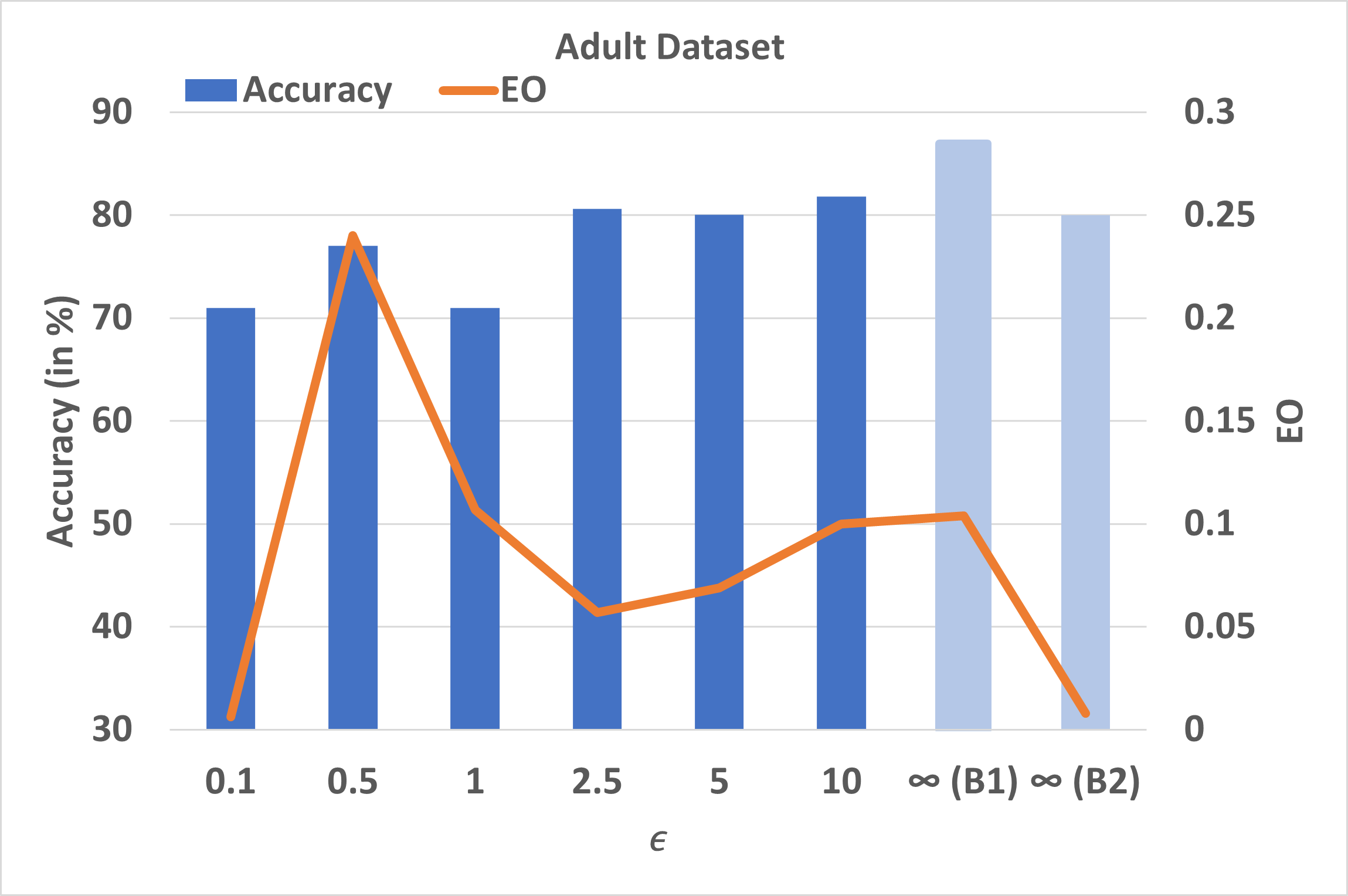}}
\caption{\label{fig:adult}Performance for the Adult dataset}
\end{figure*}
%

%
\begin{figure*}[!t]
\centering
\subfigure[Demographic Parity (DemP)]{\includegraphics[width=0.4\columnwidth]{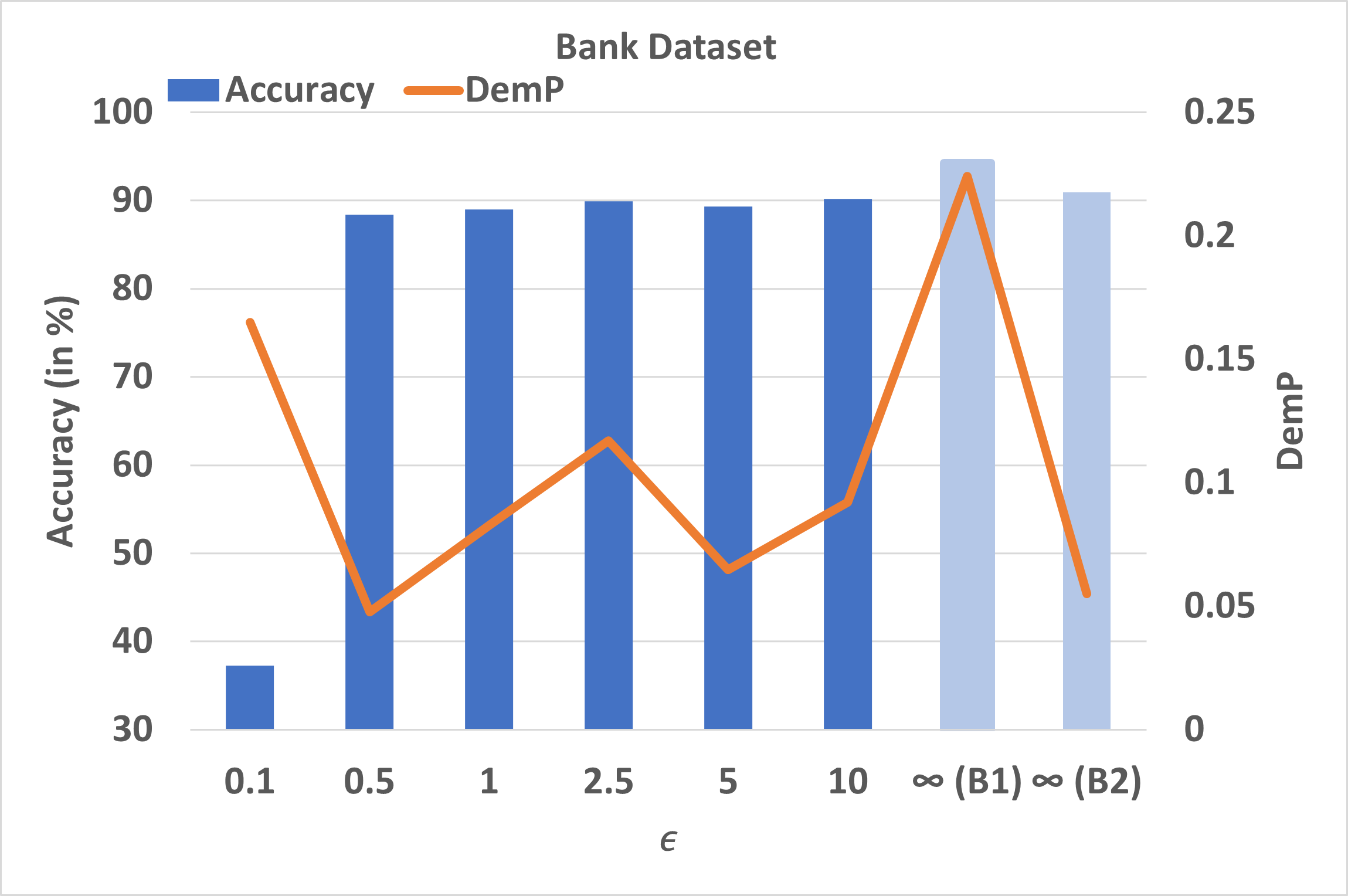}}
\smallskip
\subfigure[Equalized Odds (EO)]{\includegraphics[width=0.4\columnwidth]{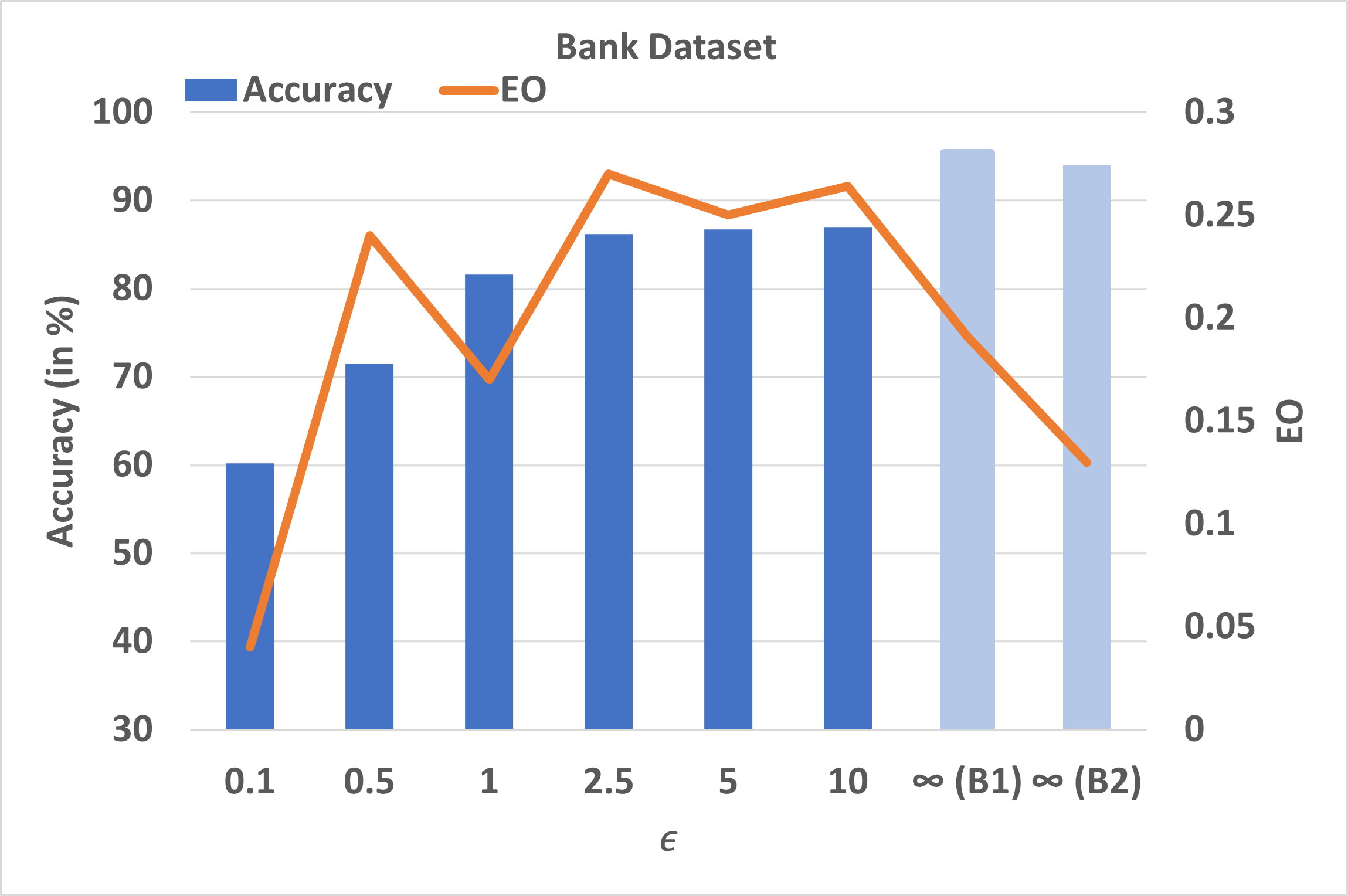}}
\caption{\label{fig:bank}Performance for the Bank dataset}
\end{figure*}
%

%
\begin{figure*}[!t]
\centering
\subfigure[Demographic Parity (DemP)]{\includegraphics[width=0.4\columnwidth]{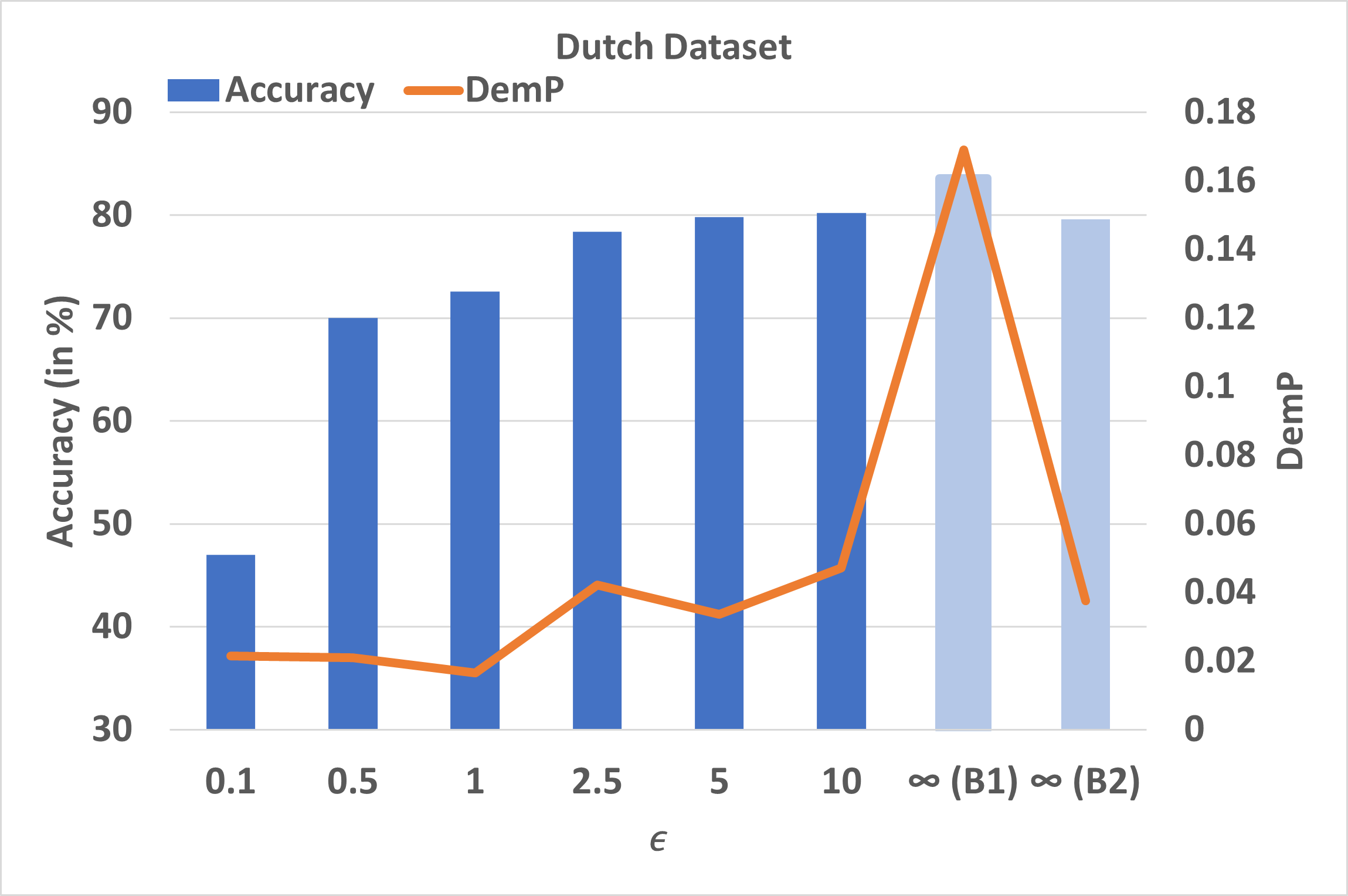}}
\smallskip
\subfigure[Equalized Odds (EO)]{\includegraphics[width=0.4\columnwidth]{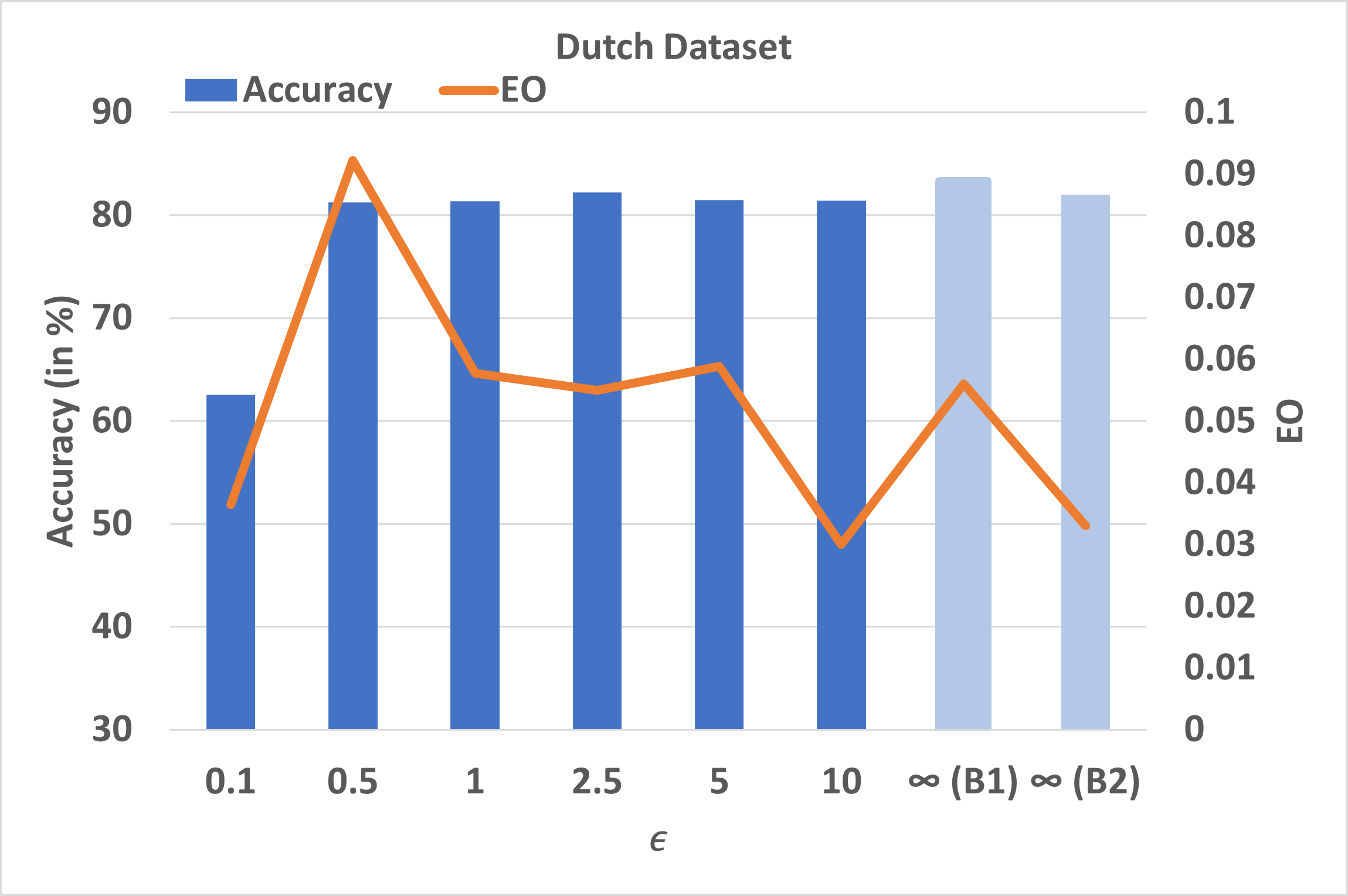}}
\caption{\label{fig:dutch}Performance for the Dutch dataset}
\end{figure*}
%

\smallskip
\noindent\textbf{Demographic Parity: Figures~\ref{fig:adult}(a), \ref{fig:bank}(a), and \ref{fig:dutch}(a).} We consider an FL setting with 5 agents for ensuring DemP. For the Adult dataset, Figure~\ref{fig:adult}(a), we find that for B1, we get an accuracy of $87\%$ and a DemP of $0.17$. We observe that a model trained with fairness constraints, i.e., for B2, has a reduced accuracy of $85\%$, but DemP reduces to $0.029$. We find similar trends in the baselines for the Bank (Figure~\ref{fig:bank}(a)) and the Dutch datasets (Figure~\ref{fig:dutch}(a)). 

Introducing privacy guarantees with FPFL, we observe a further compromise in either accuracy and fairness as compared to our baselines. In general, with increasing $\epsilon$, i.e., increasing privacy loss, there is an improvement in the trade-off of accuracy and DemP. For $\epsilon = 10$, the accuracy and DemP are similar to that in B2. While the drop in accuracy is consistent with decrease in $\epsilon$, DemP values do not always follow this trend.

\smallskip
\noindent\textbf{Equalized Odds: Figures~\ref{fig:adult}(b), \ref{fig:bank}(b), and \ref{fig:dutch}(b).}
For EO, we consider FL setting with only 2 agents. From Figure \ref{fig:adult}(b), we find in B1 the accuracy is $87\%$ for the Adult dataset with EO as $0.104$. With B2, we obtain reduced accuracy of $80\%$, but EO reduces to $0.008$. We find similar trends in the baselines for the Bank (Figure~\ref{fig:bank}(b)) and the Dutch datasets (Figure~\ref{fig:dutch}(b)). 

When we compare the FPFL training, which also guarantees privacy, we observe a trade-off in fairness and accuracy. We note that ensuring EO, especially in the Bank dataset, is very challenging. Therefore, the trade-off is not as smooth. With decrease in $\epsilon$, the accuracy decreases, but the EO values do not follow any trend. We believe this is due to the lack of distinct samples for each category after splitting the data (despite duplication) for FL. 

\smallskip
\noindent\textit{Future Work.}  Our goal is to establish the FPFL framework, where a user can customize $\lambda$ and $\sigma$ to ensure the desired performance. The framework allows the use of any fairness measure of choice by appropriately modifying the loss in Phase 1. Exploring these on other relevant datasets and exploring other fairness~\cite{li2020,du2021fairness} and privacy techniques~\cite{triastcyn2019federated} with varying number of clients is left for future work. 


\section{Conclusion}
We provide a framework that learns fair and accurate models while preserving privacy. We refer to our novel framework as FPFL (Figure~\ref{fig::FPFL-framework}). We showed that decoupling the training process into separated phases for fairness and privacy allowed us to provide a DP guarantee for the training data and sensitive attributes and reduce the number of training timesteps. We then applied FPFL on the Adult, Bank and Dutch datasets to highlight the relation between accuracy, fairness, and privacy of an FL model.

\nocite{langley00}

\bibliographystyle{splncs04}
\bibliography{ref}




\end{document}

%% file: relatedwork.tex
\begin{table}[!t]
    \centering
    \def\arraystretch{1.15}
    \begin{adjustbox}{width=0.8\textwidth}
    \begin{tabular}{c||c|c||c|c||c}

    \multirow{2}{*}{\textbf{Paper}}     &  \multicolumn{2}{c||}{\textbf{Fairness}} & \multicolumn{2}{c||}{\textbf{Privacy}} & \multirow{2}{*}{\textbf{FL}} \\
\cline{2-5}
         &  Demographic Parity & Equalized Odds & Training Data & Sensitive Attribute & \\
\hline
\cite{cummings2019compatibility,mozannar2020fair} & \xmark& \cmark & \cmark & \xmark& \xmark\\ 
\cite{tran2020differentially} & \cmark & \cmark & \xmark & \cmark & \xmark \\ 
\cite{du2021fairness} & \cmark & \xmark & \xmark & \xmark & \cmark \\ 
\cite{naseri2020toward,triastcyn2019federated,ieeefpdp}&  \xmark&  \xmark & \cmark &  \xmark& \cmark \\ 
FPFL & \cmark & \cmark & \cmark & \cmark & \cmark \\
\hline
    \end{tabular}
    \end{adjustbox}
    \caption{Comparing Existing Literature with FPFL}
    \label{tab:RW}
\end{table}

%% file: FPFL-Algo.tex
\begin{algorithm}[tb]
  \caption{Fair-SGD for an Agent $i$}\label{algo::P1}
\begin{algorithmic}
   \STATE {\bfseries Input:} Training dataset $\mathcal{X}_i=\{x_1,\dots,x_n\}$, Loss function $L_1(\cdot)$ as defined in \eqref{eqn::Phase1Loss}. Hyperparameters: learning rate $\eta, \eta'$, batch size $B$, sampling probability $q=B/|\mathcal{X}_i|$.
 \STATE {\bfseries Output:} $\phi_{(i,T_1)}$
 \STATE {\bfseries Initialization:} $\phi_{(i,0)}\leftarrow \mbox{~randomly}$, $\lambda^{max}_i \leftarrow $ max value 
 \FOR{$t\in [T_1]$}
  \STATE Take a random sample $B_t$ with probability $q$ 
 \STATE $\forall k\in B_t$:
    $
    \mathbf{g}_t(x_k) \leftarrow \nabla_{\phi_{(i,t)}} L_1(\cdot);  \mathbf{g'}_t(x_k) \leftarrow \nabla_{\lambda_{(i,t)}} L_1(\cdot)
    $
 
\STATE $\phi_{(i,t+1)}\leftarrow \phi_{(i,t)} -  \eta_t{\mathbf{g}}_t(x_k)$;
$\lambda_{(i, t+1)} \leftarrow \lambda_{(i,t)} + \eta' g'_t(x_k)$
 \ENDFOR
 \end{algorithmic}
\end{algorithm}

\begin{algorithm}[tb]
  \caption{DP-SGD for an Agent $i$~\cite{abadi2016deep}}\label{algo::P2}
\begin{algorithmic}
   \STATE {\bfseries Input:} Training dataset $\mathcal{X}_i=\{x_1,\dots,x_n\}$, Loss function $L_2(\cdot)$ as defined in \eqref{eqn::Phase2Loss}. Hyperparameters: learning rate $\eta$, standard deviation $\sigma$, batch size $B$, sampling probability $q=B/|\mathcal{X}_i|$ and clipping norm $C$.
 \STATE {\bfseries Output:} $\theta_{(i,T_2)}$
 \STATE {\bfseries Initialization:} $\theta_{(i,0)}\leftarrow \mbox{~randomly}$
 \FOR{$t\in [T_2]$}
 \STATE Take a random sample $B_t$ with probability $q$ 
 \STATE $\forall k\in B_t$:
    $
    \mathbf{g}_t(x_k) \leftarrow \nabla_{\theta_{(i,t)}} L_2(\cdot)
    $
 \STATE $\bar{\mathbf{g}}_t(x_k) \leftarrow \mathbf{g}_t(x_k)/\max\left(1, \frac{||\mathbf{g}_t(x_k)||_2}{C}\right) $;
$\Tilde{\mathbf{g}}_t(x_k) \leftarrow \frac{1}{B}
\left(\sum_i \bar{\mathbf{g}}_t(x_k) + \mathcal{N}(0,\sigma^2C^2\mathbf{I}) \right)$
\STATE $\theta_{(i,t+1)}\leftarrow \theta_{(i,t)} -  \eta_t\Tilde{\mathbf{g}}_t(x_k)$
 \ENDFOR
 \end{algorithmic}
\end{algorithm}

%% file: samplepaper.bbl
\begin{thebibliography}{10}
\providecommand{\url}[1]{\texttt{#1}}
\providecommand{\urlprefix}{URL }
\providecommand{\doi}[1]{https://doi.org/#1}

\bibitem{abadi2016deep}
Abadi, M., Chu, A., Goodfellow, I., McMahan, H.B., Mironov, I., Talwar, K.,
  Zhang, L.: Deep learning with differential privacy. In: Proceedings of the
  2016 ACM SIGSAC conference on computer and communications security. pp.
  308--318 (2016)

\bibitem{agarwal18}
Agarwal, A., Beygelzimer, A., Dudik, M., Langford, J., Wallach, H.: A
  reductions approach to fair classification. In: Dy, J., Krause, A. (eds.)
  Proceedings of the 35th International Conference on Machine Learning.
  Proceedings of Machine Learning Research, vol.~80, pp. 60--69. PMLR,
  Stockholmsmässan, Stockholm Sweden (10--15 Jul 2018),
  \url{http://proceedings.mlr.press/v80/agarwal18a.html}

\bibitem{bagdasaryan2019differential}
Bagdasaryan, E., Poursaeed, O., Shmatikov, V.: Differential privacy has
  disparate impact on model accuracy. Advances in Neural Information Processing
  Systems  \textbf{32},  15479--15488 (2019)

\bibitem{barocas16}
Barocas, S., Selbst, A.D.: Big data's disparate impact. Cal. L. Rev.
  \textbf{104}, ~671 (2016)

\bibitem{berk}
Berk, R., Heidari, H., Jabbari, S., Kearns, M., Roth, A.: Fairness in criminal
  justice risk assessments: The state of the art. Sociological Methods \&
  Research p. 0049124118782533 (2018)

\bibitem{zafar17}
{Bilal Zafar}, M., {Valera}, I., {Gomez Rodriguez}, M., {Gummadi}, K.P.:
  {Fairness Constraints: Mechanisms for Fair Classification}. ArXiv e-prints
  (Jul 2015)

\bibitem{chouldechova17}
Chouldechova, A.: Fair prediction with disparate impact: A study of bias in
  recidivism prediction instruments. Big data  \textbf{5 2},  153--163 (2017)

\bibitem{cummings2019compatibility}
Cummings, R., Gupta, V., Kimpara, D., Morgenstern, J.: On the compatibility of
  privacy and fairness. In: Adjunct Publication of the 27th Conference on User
  Modeling, Adaptation and Personalization. pp. 309--315 (2019)

\bibitem{du2021fairness}
Du, W., Xu, D., Wu, X., Tong, H.: Fairness-aware agnostic federated learning.
  In: Proceedings of the 2021 SIAM International Conference on Data Mining
  (SDM). pp. 181--189. SIAM (2021)

\bibitem{dwork2012fairness}
Dwork, C., Hardt, M., Pitassi, T., Reingold, O., Zemel, R.: Fairness through
  awareness. In: Proceedings of the 3rd innovations in theoretical computer
  science conference. pp. 214--226 (2012)

\bibitem{dwork2014algorithmic}
Dwork, C., Roth, A., et~al.: The algorithmic foundations of differential
  privacy. Foundations and Trends in Theoretical Computer Science
  \textbf{9}(3-4),  211--407 (2014)

\bibitem{fang2021privacy}
Fang, H., Qian, Q.: Privacy preserving machine learning with homomorphic
  encryption and federated learning. Future Internet  \textbf{13}(4), ~94
  (2021)

\bibitem{fredrikson2015model}
Fredrikson, M., Jha, S., Ristenpart, T.: Model inversion attacks that exploit
  confidence information and basic countermeasures. In: Proceedings of the 22nd
  ACM SIGSAC Conference on Computer and Communications Security. pp. 1322--1333
  (2015)

\bibitem{hardt16}
Hardt, M., Price, E., Srebro, N.: Equality of opportunity in supervised
  learning. In: NIPS (2016)

\bibitem{hinton2015distilling}
Hinton, G., Vinyals, O., Dean, J.: Distilling the knowledge in a neural
  network. arXiv preprint arXiv:1503.02531  (2015)

\bibitem{li2020}
Li, T., Sanjabi, M., Beirami, A., Smith, V.: Fair resource allocation in
  federated learning. In: 8th International Conference on Learning
  Representations, {ICLR} 2020, Addis Ababa, Ethiopia, April 26-30, 2020.
  OpenReview.net (2020), \url{https://openreview.net/forum?id=ByexElSYDr}

\bibitem{madras18}
Madras, D., Creager, E., Pitassi, T., Zemel, R.S.: Learning adversarially fair
  and transferable representations. In: Proceedings of the 35th International
  Conference on Machine Learning, {ICML} 2018, Stockholmsm{\"{a}}ssan,
  Stockholm, Sweden, July 10-15, 2018. pp. 3381--3390 (2018),
  \url{http://proceedings.mlr.press/v80/madras18a.html}

\bibitem{mcmahan2017communication}
McMahan, B., Moore, E., Ramage, D., Hampson, S., y~Arcas, B.A.:
  Communication-efficient learning of deep networks from decentralized data.
  In: Artificial Intelligence and Statistics. pp. 1273--1282. PMLR (2017)

\bibitem{mohassel2017secureml}
Mohassel, P., Zhang, Y.: Secureml: A system for scalable privacy-preserving
  machine learning. In: 2017 IEEE Symposium on Security and Privacy (SP). pp.
  19--38. IEEE (2017)

\bibitem{mothukuri2021survey}
Mothukuri, V., Parizi, R.M., Pouriyeh, S., Huang, Y., Dehghantanha, A.,
  Srivastava, G.: A survey on security and privacy of federated learning.
  Future Generation Computer Systems  \textbf{115},  619--640 (2021)

\bibitem{mozannar2020fair}
Mozannar, H., Ohannessian, M., Srebro, N.: Fair learning with private
  demographic data. In: International Conference on Machine Learning. pp.
  7066--7075. PMLR (2020)

\bibitem{naseri2020toward}
Naseri, M., Hayes, J., De~Cristofaro, E.: Toward robustness and privacy in
  federated learning: Experimenting with local and central differential
  privacy. arXiv preprint arXiv:2009.03561  (2020)

\bibitem{manisha2018fnnc}
Padala, M., Gujar, S.: Fnnc: Achieving fairness through neural networks. In:
  Bessiere, C. (ed.) Proceedings of the Twenty-Ninth International Joint
  Conference on Artificial Intelligence, {IJCAI-20}. pp. 2277--2283.
  International Joint Conferences on Artificial Intelligence Organization (7
  2020). \doi{10.24963/ijcai.2020/315},
  \url{https://doi.org/10.24963/ijcai.2020/315}, main track

\bibitem{pathak2010multiparty}
Pathak, M.A., Rane, S., Raj, B.: Multiparty differential privacy via
  aggregation of locally trained classifiers. In: NIPS. pp. 1876--1884.
  Citeseer (2010)

\bibitem{shokri2015privacy}
Shokri, R., Shmatikov, V.: Privacy-preserving deep learning. In: Proceedings of
  the 22nd ACM SIGSAC conference on computer and communications security. pp.
  1310--1321 (2015)

\bibitem{tran2020differentially}
Tran, C., Fioretto, F., Van~Hentenryck, P.: Differentially private and fair
  deep learning: A lagrangian dual approach. arXiv preprint arXiv:2009.12562
  (2020)

\bibitem{triastcyn2019federated}
Triastcyn, A., Faltings, B.: Federated learning with bayesian differential
  privacy. In: 2019 IEEE International Conference on Big Data (Big Data). pp.
  2587--2596. IEEE (2019)

\bibitem{wahab2021federated}
Wahab, O.A., Mourad, A., Otrok, H., Taleb, T.: Federated machine learning:
  Survey, multi-level classification, desirable criteria and future directions
  in communication and networking systems. IEEE Communications Surveys \&
  Tutorials  (2021)

\bibitem{ieeefpdp}
Wei, K., Li, J., Ding, M., Ma, C., Yang, H.H., Farokhi, F., Jin, S., Quek,
  T.Q.S., Poor, H.V.: Federated learning with differential privacy: Algorithms
  and performance analysis. IEEE Transactions on Information Forensics and
  Security  \textbf{15},  3454--3469 (2020). \doi{10.1109/TIFS.2020.2988575}

\bibitem{yang2019federated}
Yang, Q., Liu, Y., Chen, T., Tong, Y.: Federated machine learning: Concept and
  applications. ACM Transactions on Intelligent Systems and Technology (TIST)
  \textbf{10}(2),  1--19 (2019)

\bibitem{zhang2020batchcrypt}
Zhang, C., Li, S., Xia, J., Wang, W., Yan, F., Liu, Y.: Batchcrypt: Efficient
  homomorphic encryption for cross-silo federated learning. In: 2020
  $\{$USENIX$\}$ Annual Technical Conference ($\{$USENIX$\}$$\{$ATC$\}$ 20).
  pp. 493--506 (2020)

\bibitem{dutch}
Žliobaite, I., Kamiran, F., Calders, T.: Handling conditional discrimination.
  In: 2011 IEEE 11th International Conference on Data Mining. pp. 992--1001
  (2011). \doi{10.1109/ICDM.2011.72}

\end{thebibliography}
